\pdfoutput=1

\documentclass[11pt]{article}

\usepackage[preprint]{acl}

\usepackage{times}
\usepackage{latexsym}
\usepackage{booktabs}

\usepackage{subcaption}

\usepackage[T1]{fontenc}
\usepackage[utf8]{inputenc}

\usepackage{microtype}

\usepackage{inconsolata}

\usepackage{graphicx}

\usepackage{xcolor}
\usepackage{authblk}

\title{Differentially-private text generation degrades output language quality}
\author[]{\textbf{Erion {\c C}ano}}
\author[]{\textbf{Ivan Habernal}}
\affil[]{Trustworthy Human Language Technologies \\
Research Center Trustworthy Data Science and Security of the University Alliance Ruhr, \\
Faculty of Computer Science, Ruhr University Bochum \\
\{\texttt{erion.cano, ivan.habernal}\}@ruhr-uni-bochum.de \\
\href{http://www.trusthlt.org/}{www.trusthlt.org}
}

\begin{document}

\maketitle

\begin{abstract}
Ensuring user privacy by synthesizing data from large language models (LLMs) tuned under differential privacy (DP) has become popular recently. However, the impact of DP fine-tuned LLMs on the quality of the language and the utility of the texts they produce has not been investigated. In this work, we tune five LLMs with three corpora under four levels of privacy and assess the length, the grammatical correctness, and the lexical diversity of the text outputs they produce. We also probe the utility of the synthetic outputs in downstream classification tasks such as book genre recognition based on book descriptions and cause of death recognition based on verbal autopsies. The results indicate that LLMs tuned under stronger privacy constrains produce texts that are shorter by at least 77\,\%, that are less grammatically correct by at least 9\,\%, and are less diverse by at least 10\,\% in bi-gram diversity. Furthermore, the accuracy they reach in downstream classification tasks decreases, which might be detrimental to the usefulness of the generated synthetic data.
\end{abstract}

\section{Introduction}

Synthetic text generation helps to address certain problems like data scarcity, since LLMs of today are exhausting the available human written text collections \cite{villalobos2024rundatalimitsllm}. In fact, Microsoft reports that synthetic texts constitute the bulk of the data in its latest Phi-4 model \cite{abdin2024phi4technicalreport}. 

Using synthetic texts may also help to alleviate privacy concerns.
One method to generate private synthetic texts is by tuning LLMs with DP, an increasingly popular framework that provides formal guarantees on the level of privacy to expect under certain conditions \cite{10.1007/11787006_1}. 
In fact, several studies are proposing LLMs tuned with DP for creating private synthetic texts \cite{yue_synthetic_2023,yu_differentially_2022,ochs2024privatesynthetictextgeneration}. There is still one unaddressed problem: what happens with the language of the synthetic texts that are generated using DP? 

\citet{guo-etal-2024-curious} explore the lexical diversity of texts, reporting a decline when synthetic outputs are used iteratively to fine-tune LLMs which then produce more synthetic texts. However, they do not examine the role of DP and the respective language implications. Moreover, existing works on DP have used synthetic data only to train downstream classifiers. To the best of our knowledge, there is still no work to directly observe the influence of DP on the linguistic properties of the synthetic texts obtained from LLMs. 

To address this, we fine-tune several LLMs under certain DP constrains and examine the length, the lexical diversity, and the grammatical correctness of the produced text sequences, as well as the accuracy they reach in downstream classification tasks. We work with different public and private text collections and answer the following research questions:
\begin{description}
\item[\hypertarget{RQ1}{\textbf{RQ1:}}]Which aspects of language quality are impacted by DP fine-tuning and how? %
\item[\hypertarget{RQ2}{\textbf{RQ2:}}] What role does the privacy budget $\varepsilon$ play?
\item[\hypertarget{RQ3}{\textbf{RQ3:}}] What is the impact DP tuning on data utility in downstream classification tasks?
\end{description}

We experimented with five open LLMs (see \S\,\ref{ssec:llms}) and 
three corpora (see \S\,\ref{ssec:corpora}), assessing output length, grammatical correctness, and lexical diversity. 
We also investigate the utility of the outputs in downstream classification tasks. More specifically, we frame and run two such tasks: book genre recognition based on book descriptions and cause of death recognition based on verbal autopsies.

Based on our results, the average length of the synthetic outputs gets significantly smaller under stronger privacy constrains (smaller $\varepsilon$), losing 77\,\% of its original length in the best case. Moreover, there is a noticeable loss  by at least 9\,\% in grammatical correctness and at least 10\,\% in bi-gram diversity. As for the classification utility, we noticed a slight negative impact on the cause of death recognition task and a strong negative impact on book genre recognition. 

\section{Related Work} %
\label{sec:related}

\subsection{DP for Text Privacy} %
\label{ssec:dptextgen}

The current demand for higher data efficiency \cite{liu2025shiftingaiefficiencymodelcentric,cano-bojar-2019-efficiency} and the rising concerns related to user privacy \cite{WU2024102,ALKAMLI2024e39087} have boosted research interest in the adoption of DP when working with texts and LLMs. The DP framework provides formal guarantees about the privacy of each data record in a dataset by using randomization noise \cite{10.1561/0400000042}. 
Several authors are adopting it to protect user privacy. 
\citet{mattern_differentially_2022} propose a DP mechanism for mirroring private sensitive data into synthetic ones. They utilize GPT-2 \cite{radford2019language} tuned with a DP algorithm to protect the content of their training data and sample synthetic and anonymous texts from it.

\citet{yu_differentially_2022} propose a framework that tunes LLMs with private data, freezing the original parameters and adding extra ones. The trained extra parameters can become public and can be attached as plug-in to the original model, ensuring the privacy of the data used to produce them. 
Furthermore, \citet{yue_synthetic_2023} use DP-SGD \cite{6736861} to tune LLMs with privacy guarantees and sample synthetic data. They compare the topical similarity of the synthetic texts with the original ones and report that it increases by using larger LLMs. 

\citet{ochs2024privatesynthetictextgeneration} compare the quality of DP synthetic texts from LLMs with those obtained from diffusion models \cite{gong_diffuseq_2023}. They conclude that the later provide good results, but do not surpass the former.
Finally, \citet{ijcai2024p735} control the alignment between synthetic texts and the original sensitive texts with an iterative algorithm. They report that the synthetic samples closely match the utility of the private ones.

All these works provide several ways of fine-tuning LLMs with DP and evaluate the outputs w.r.t their utility and privacy. However, none of them loops into the language aspects of what comes out of the DP-tuning process. 

\subsection{Text Quality Evaluation}
\label{ssec:qualityeval}

An important aspect of text quality evaluation is grammatical correctness which relates to tasks like machine translation, text summarization, question answering, etc. It has been traditionally assessed using reference-based metrics (comparing obtained outputs against gold standard references) such as MaxMatch \cite{dahlmeier-ng-2012-better} and GLEU \cite{napoles_ground_2015}. On the other hand, GBMs (Grammaticality-Based Metrics) do not require gold standard references and there is evidence that they correlate very well with human judgement \cite{napoles_theres_2016}. The idea behind GBMs is to count grammatical errors using error detection tools 
and compute the grammatical correctness score by normalizing the count per unit length. In other words, we compute:
\begin{equation}
\mathrm{GC} = 1 - \frac{{\scriptstyle\#}\, \mathrm{errors}}{{\scriptstyle\#}\,\mathrm{tokens}}
\label{eq:gc}
\end{equation}
and obtain a value from 0 to 1. 
Assessing lexical diversity is also common, since it provides insights about language richness. One way is to measure \textit{distinct-n} which is the ratio between the unique \textit{n-grams} and the total \textit{n-grams} in a text \cite{li-etal-2016-diversity}. Another metric is Self-BLEU \cite{10.1145/3209978.3210080} that compares the sentences pairwise using BLEU \cite{papineni-etal-2002-bleu} and returns their average similarity. 
HS (Homogenization Score) is similar to Self-BLEU, but uses Rouge-L \cite{lin-2004-rouge} for the pairwise comparisons. Finally, lexical diversity can also be measured using compression algorithms. One such approach is proposed by \citet{shaib_standardizing_2024} who compute the compression score as the ratio between the size of the output and its compressed version using gzip algorithm.

\section{Experimental Setup}
\label{sec:expsetup}

\begin{figure*}[h]
    \centering
    \includegraphics[scale=0.38]{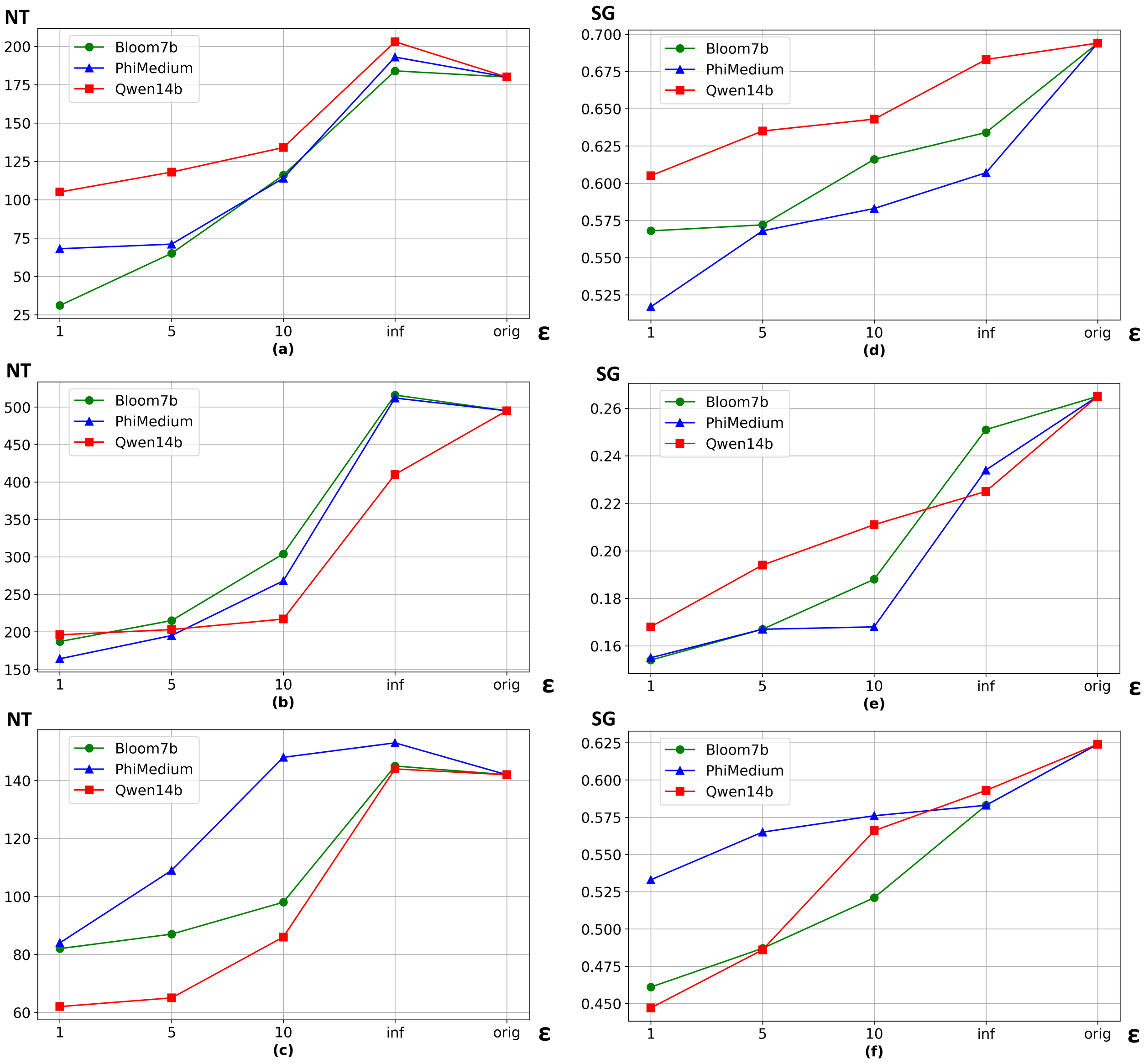}
    \caption{Number of tokens (NT) for each privacy level and dataset (Autopsy on (a), Booksum on (b), and Suicide on (c)) on the left, and sentence-level grammaticality scores (SG) for each privacy level and dataset (Autopsy on (d), Booksum on (e), and Suicide on (f)) on the right.}
    \label{fig:allntsg}
\end{figure*}

\begin{figure*}[h]
    \centering
    \includegraphics[scale=0.38]{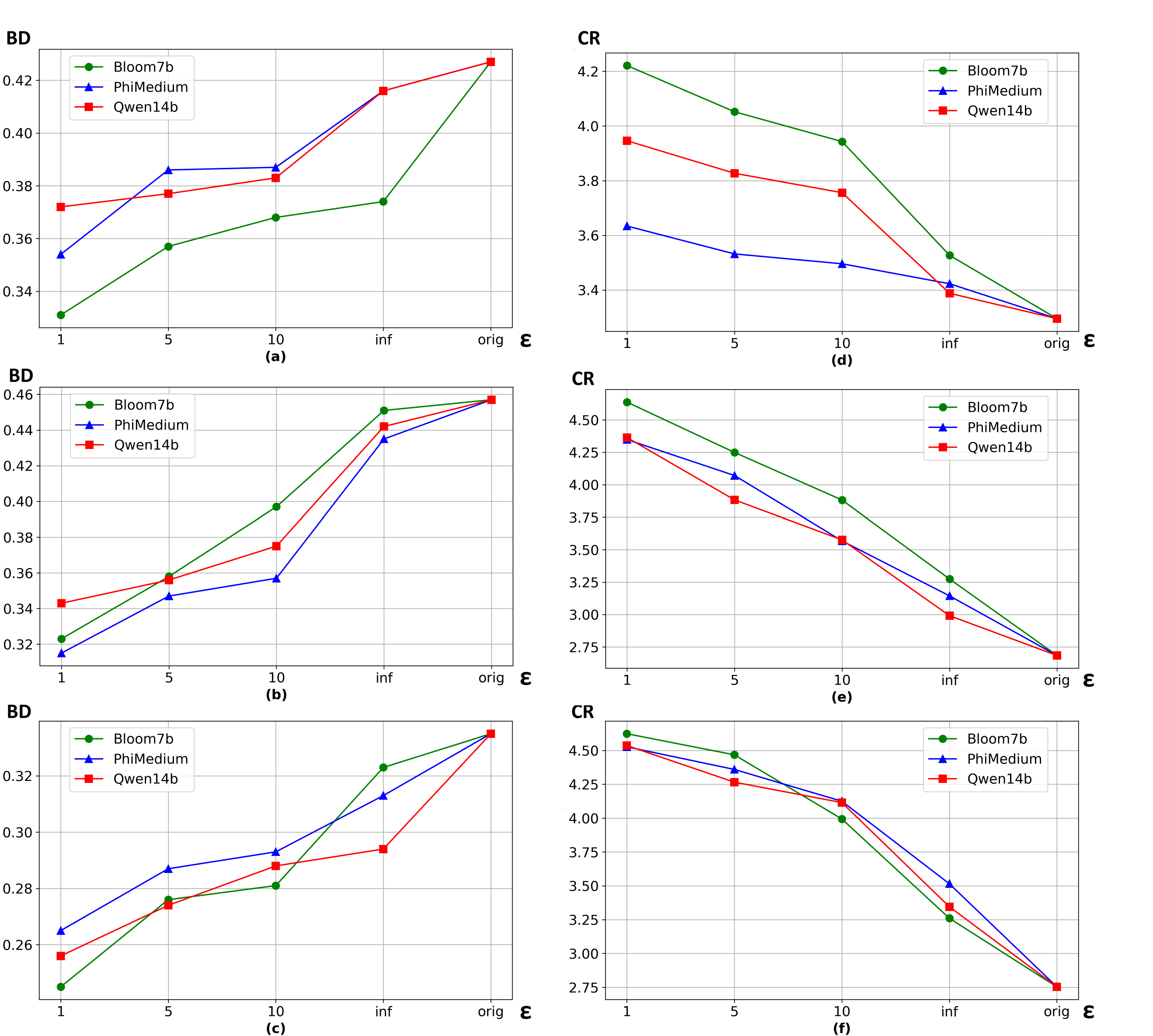}
    \caption{Bigram diversity (BD) for each privacy level and dataset (Autopsy on (a), Booksum on (b), and Suicide on (c)) on the left, and compression ratio (CR) for each privacy level and dataset (Autopsy on (d), Booksum on (e), and Suicide on (f)) on the right.}
    \label{fig:allbdcr}
\end{figure*}

\subsection{Language Models} \label{ssec:llms}

Bloom is a suit of fully open LLMs (weights, code, and data) released for research purposes \cite{2022arXiv221105100W}. Its biggest variant with 176 billion parameters was the first to compare against GPT-3 \cite{NEURIPS2020_1457c0d6}. 
Due to technical limitations (see \S\,\ref{sec:limitations}), the biggest variant we could use is Bloom7b\footnote{\url{https://huggingface.co/bigscience/bloom-7b1}} with 7 billion parameters.
Phi suite of LLMs represent models that are relatively small in size, but compare in performance with LLMs at least five times bigger. Phi-4 mini has only 3.8 billion parameters and was released recently \cite{microsoft2025phi4minitechnicalreportcompact}. Another model is the the medium variant of 14 billion parameters\footnote{\url{https://huggingface.co/microsoft/phi-4}} which is also recent \cite{abdin2024phi4technicalreport}. They are both decoder-only architectures like most of the LLMs, but their extra value comes in the quality of the data they have been pretrained with. It is a combination of highly curated web content, public books, and high-quality synthetic texts.   

Qwen-2.5 is an instance of the Qwen suite \cite{qwen2025qwen25technicalreport}. These models support many natural languages, are pretrained with a huge amount of texts (about 18 trillion tokens), and can capture a very long context window. In this work, we experiment with Qwen-2.5-7B\footnote{\url{https://huggingface.co/Qwen/Qwen2.5-7B}} and Qwen-2.5-14B.\footnote{\url{https://huggingface.co/Qwen/Qwen2.5-14B}}

We use $(\varepsilon, \delta)$ DP and fine-tuned the five models with each of the three datasets on four privacy levels: $\varepsilon = \infty$ for synthetic texts with no privacy guarantees at all, $\varepsilon = 10$ for weak privacy, $\varepsilon = 5$ for moderate privacy, and $\varepsilon = 1$ for strong privacy. This is similar to \citet{yue_synthetic_2023} who also use different levels of $\varepsilon$. 

As for the $\delta$, we compute and utilize three values (each value for all experiments on each dataset) using the formula $\delta_d = 1 / (n_d \cdot 10)$, where $n_d$ represents the number of training samples of the dataset. This results in $\delta$ values of $1.67 \cdot 10^{-5}$, $10^{-5}$, and $7.05 \cdot 10^{-6}$ for Autopsy, Booksum, and Suicide respectively.

Finally, we use the recently proposed encoder-only ModernBERT \citep{warner_smarter_2024} model as classifier and fine-tune it with synthetic texts of book descriptions and verbal autopsies.

\subsection{Text Corpora} \label{ssec:corpora}
Choosing relevant data is important for running realistic and real-world privacy related experiments. One desiderata is to work with texts that convey sensitive information about delicate topics, potentially revealing personal data, referring to minority groups, etc. Protecting leakage of such information is one of the main reasons for developing data privacy methods.   
Another desiderata is to work with texts that reflect diverse language styles coming from various contexts. This way we can better detect and assess stylistic changes in language.    

Based on the above considerations, we decided to work with two private corpora, namely Verbal Autopsies (referred to as \textit{Autopsy}) and Suicide Detection (referred to as \textit{Suicide}), contrary to other similar studies that utilize public data only \cite{amin_private_2024,yue_synthetic_2023}. To cope with the second desiderata, we extend the experimental setup including a public collection also, the CMU Book Summary Dataset referred to as \textit{Booksum}. 

Autopsy is a corpus of 11\,978 verbal autopsy interviews, each conducted by a field interviewer who asked an individual familiar with the deceased about their illness or death cause \cite{flaxman_-identified_2018}. A semi-structured questionnaire was used to collect information about the symptoms and other possible risk factors. Sensitive data like names, locations, or dates are suppressed with placeholders. The dataset is not publicly available, but can be obtained upon request (see \S\,\ref{sec:ethicalconsiderations}).

Suicide is a collection of posts related to mental disorders and suicide intentions, collected from Reddit subreddits \cite{ji_suicidal_2022}. The data have been categorized in five classes, namely `depression', `suicide', `anxiety', `offmychest', and `bipolar', based on the specific subreddit they were obtained from. This dataset is also not publicly available, but can be obtained upon request.

Finally, Booksum corpus contains publicly available book summaries \citet{DBLP:journals/corr/abs-1305-1319}. It comprises 16\,559 plot summaries written by readers of those books. Each record contains additional book attributes like title, author, year of publication, and genre. The full specifications of the data are shown in Appendix~\ref{sec:corporastats}.

\subsection{Evaluation Metrics} \label{ssec:evalmetrics}

We probed three aspects of the produced texts: sample length, grammatical correctness, and lexical diversity. 
Output sample length is important, since longer texts usually carry more information. We assess the average output length in number of characters and number of tokens. 

We also evaluate grammatical correctness using Equation~\ref{eq:gc}, computing WG (Word Grammaticality) and SG (Sentence Grammaticality). For the WG, we do a length normalization by dividing the number of errors in the sample with its average token length. Next, we compute and report the average WG of all the dataset samples. For the SG, we normalize by dividing the errors of each sentence in the sample with the length of that sentence, and then compute the per-sentence average. High values of WG and SG indicate a high grammatical correctness.

To measure lexical diversity, we compute UD (Unigram Diversity) also known as Distinct-1, as well as BD (Bigram Diversity) or Distinct-2. For longer overlaps, we measure HS (Homogenization Score) with ROUGE-L as defined and implemented by \citet{shaib_standardizing_2024}. Finally, we also compute the CR (Compression Ratio) using the gzip algorithm.
For the evaluation of the data utility in classification tasks, we assess accuracy and F$_1$ scores. They are computed and reported for each of the  combinations between the datasets, the models, and the privacy levels. 

\section{Results} \label{sec:results}

The obtained synthetic texts were readable, with occasional hallucinations appearing as out of context sentences. Examples with the utilized prompts and the respective outputs are shown in Appendix~\ref{sec:outputexamples}. In the following sections, we discuss the variations of the output length (NT scores), grammatical correctness (SG scores), and lexical diversity (BD and CR scores) for Bloom7b, PhiMed, and Qwen14b on each of the three corpora. We start with the scores for $\varepsilon = 1$ which represents the strongest privacy, and then for $\varepsilon = 5$, $\varepsilon = 10$, and finally $\varepsilon = \infty$ which refers to the synthetic data generated without DP. The scores of the other four NC, WG, UD, and HS metrics and those obtained using the other two models PhiMini and Qwen7b reveal the same trends and are shown in Appendix~\ref{sec:fullscores}.  

\subsection{Output Length} \label{ssec:outlength}

Output length w.r.t the privacy budget obtained on the Autopsy data is depicted in Figure~\ref{fig:allntsg} (a). For $\varepsilon = 1$, synthetic autopsy texts have an average length of 31 tokens on Bloom7b, 68 tokens on PhiMed, and 105 tokens on Qwen14b. They steadily grow and reaches peaks of 184, 193, and 203 tokens for no privacy ($\varepsilon = \infty$ depicted \emph{inf} in the graphics). These lengths are slightly higher than that of the original natural autopsies (\emph{orig} in the graphics) which is 180 tokens. Overall, we observe a high sensitivity of the verbal autopsy lengths, with regression factors (from the no privacy configuration to the $\varepsilon = 1$) of 5.94, 2.83, and 1.93, on Bloom7b, PhiMed, and Qwen14b respectively. 

The book summaries are on average longer than the verbal autopsies and suicide messages. Output length variations w.r.t the privacy budget obtained on the Booksum data are shown in Figure~\ref{fig:allntsg} (b). They start at 187, 164, and 196 for $\varepsilon = 1$, and peak at 516, 512, and 410 for $\varepsilon = \infty$. Meanwhile, the average token length on the original Booksum data is 495. Same as in the case of Autopsy, Booksum sample lengths are highly sensitive to the privacy level, with regression factors 2.76, 3.12, and 2.1. 

Output lengths w.r.t the privacy budget obtained on Suicide data are depicted in Figure~\ref{fig:allntsg} (c). Suicide posts are the shortest, starting at 82, 84, and 62 tokens for the strongest level of privacy. They grow steadily and reach 145, 153, and 144 tokens for no privacy, with shrinking factors 1.77, 1.82, and 2.32. The average NT for the original Suicide data is 142. 

Overall, we observe that the output length of the DP-synthetic texts is highly sensitive to the privacy level, undergoing reductions of factors ranging from 1.77 (77\,\% decrease) to 5.94 (494\,\% decrease). This phenomenon is also reported by \citet{yue_synthetic_2023}.

\subsection{Grammatical Correctness} \label{ssec:grammcorr}

Sentence-level grammatical correctness w.r.t the privacy budget obtained on the Autopsy data is depicted in Figure~\ref{fig:allntsg} (d). The models start at low scores of 0.56, 0.52, and 0.6, but jump to 0.63, 0.6, and 0.68. Meanwhile, the grammatical correctness score of the original natural texts is 0.69. We see that the regression factors between no privacy and $\varepsilon = 1$ privacy levels are 1.13, 1.15, and 1.13, which reveals a low sensitivity. We also observe that the word-level grammatical correctness follows a very similar trend (see WG scores in Tables~\ref{tab:bloomautopsy} - \ref{tab:phimediumsuicide}). 

The same observation on the Booksum data is depicted in Figure~\ref{fig:allntsg} (e). SG scores reach up to 0.15, 0.16, and 0.17 for $\varepsilon = 1$ and peak at 0.25, 0.23, and 0.23 when no DP is applied. The average SG score on the original Booksum samples is 0.26. We see that the regression factors are 1.67, 1.44, and 1.35. 

Finally, sentence-level grammatical correctness w.r.t the privacy budget obtained on the Suicide data is depicted in Figure~\ref{fig:allntsg} (f). As we can see, SG scores start at 0.46, 0.53, and 0.44. They peak at 0.54, 0.58, and 0.59 for Bloom7b, PhiMed, and Qwen14b respectively. Meanwhile, the average SG score on the original Suicide data is 0.62. The regression factors are 1.17, 1.09, and 1.34.  

Overall, we observe that the sensitivity of grammatical correctness w.r.t the privacy level is low on the Autopsy and Suicide data, but moderate on the Booksum data. The regression factors range from 1.09 (9\,\% decrease) to 1.67 (67\,\% decrease). \citet{ngong-etal-2025-differentially} observed a similar detrimental effect of DP in language and report various types of grammar errors induced by DP fine-tuning, such as word spelling mistakes, missing punctuation, etc. Furthermore, \citet{mattern_limits_2022} compared different word-level DP methods and concluded that word replacement mechanisms proposed in some DP applications tend to change the word types and introduce grammatical errors. Unfortunately, they do not provide a direct and quantitative assessment of the grammatical correctness deterioration in the DP implementations they examine. 

\subsection{Lexical Diversity} \label{ssec:lexdiv}

High lexical diversity is a desired characteristic, as it usually relates to a richer language. we assessed it by measuring the UD, BD, HS, and CR scores on each DP-synthetic output. The variation of BD scores w.r.t the privacy budget obtained on the Autopsy data is depicted in Figure~\ref{fig:allbdcr} (a). We obtained low BD scores of 0.33, 0.35, and 0.37 for $\varepsilon = 1$ on Bloom7b, PhiMed, and Qwen14b respectively. The scores reached peaks of 0.37, 0.41, and again 0.41 for $\varepsilon = \infty$. They are very close to the BD score on the original Autopsy data which is 0.42. The regression factors between the strong privacy and the zero privacy configurations are 1.12, 1.17, and 1.1. Uni-gram diversity follows a similar trend for each model. 

The situation appears very similar on the Booksum data, Figure~\ref{fig:allbdcr} (b). BD scores start at 0.32, 0.31, and 0.34 for Bloom7b, PhiMed, and Qwen14b respectively. They peak at 0.45, 0.43, and 0.44 for the synthetic data with no DP applied. The respective regression factors are 1.4, 1.39, and 1.29.

The BD scores on Suicide data shown in Figure~\ref{fig:allbdcr} (c) are even lower, starting at 0.24, 0.26, and 0.25 for $\varepsilon = 1$. When no DP is applied, they jump to 0.32, 0.31, and 0.29 for each of the three models respectively. The regression scores are 1.33, 1.19, and 1.16.

Unlike the rest of the metrics mentioned above, CR shows an increasing trend with stronger privacy constrains. However, it is worth noting that a high compression ratio implies a low lexical diversity and vice versa. The CR score trends w.r.t the privacy budget obtained on the Autopsy data are depicted in Figure~\ref{fig:allbdcr} (d). They start at 4.22, 3.63, and 3.95 for $\varepsilon = 1$ and gradually go down to 3.53, 3.42, and 3.39 for $\varepsilon = \infty$, with shrinking factors 1.4, 1.08, and 1.17. The CR score obtained on the original Autopsy data is 3.29. 

The CR scores on Booksum data start at slightly higher values of 4.63, 4.34, and 4.36 as shown in Figure~\ref{fig:allbdcr} (e), but fall to 3.27, 3.14, and 2.99 as the privacy level decreases. The respective regression factors are 1.41, 1.38, and 1.46, and the CR score for the original Booksum data is 2.68. 

Finally, the compression ratio scores on Suicide data shown in Figure~\ref{fig:allbdcr} (f) fall from 4.62, 4.52, and 4.53 to 3.26, 3.51, and 3.34 respectively. They exhibit regression factors of 1.42, 1.29 and 1.36. 
The results show moderate sensitivity of the lexical diversity w.r.t the different privacy levels. The respective regression factors range between 1.1 (10\,\% decrease) and 1.4 (40\,\% decrease) for BD, and progression factors between 1.08 (8\,\% increase) and 1.46 (46\,\% increase) for CR. Related studies \cite{guo_curious_2024} show a similar trend, indicating that synthetic text generation does indeed drain the linguistic richness in texts.   

\subsection{Classification Utility} \label{ssec:resultsclass}

\begin{figure*}[h]
    \centering
    \includegraphics[scale=0.38]{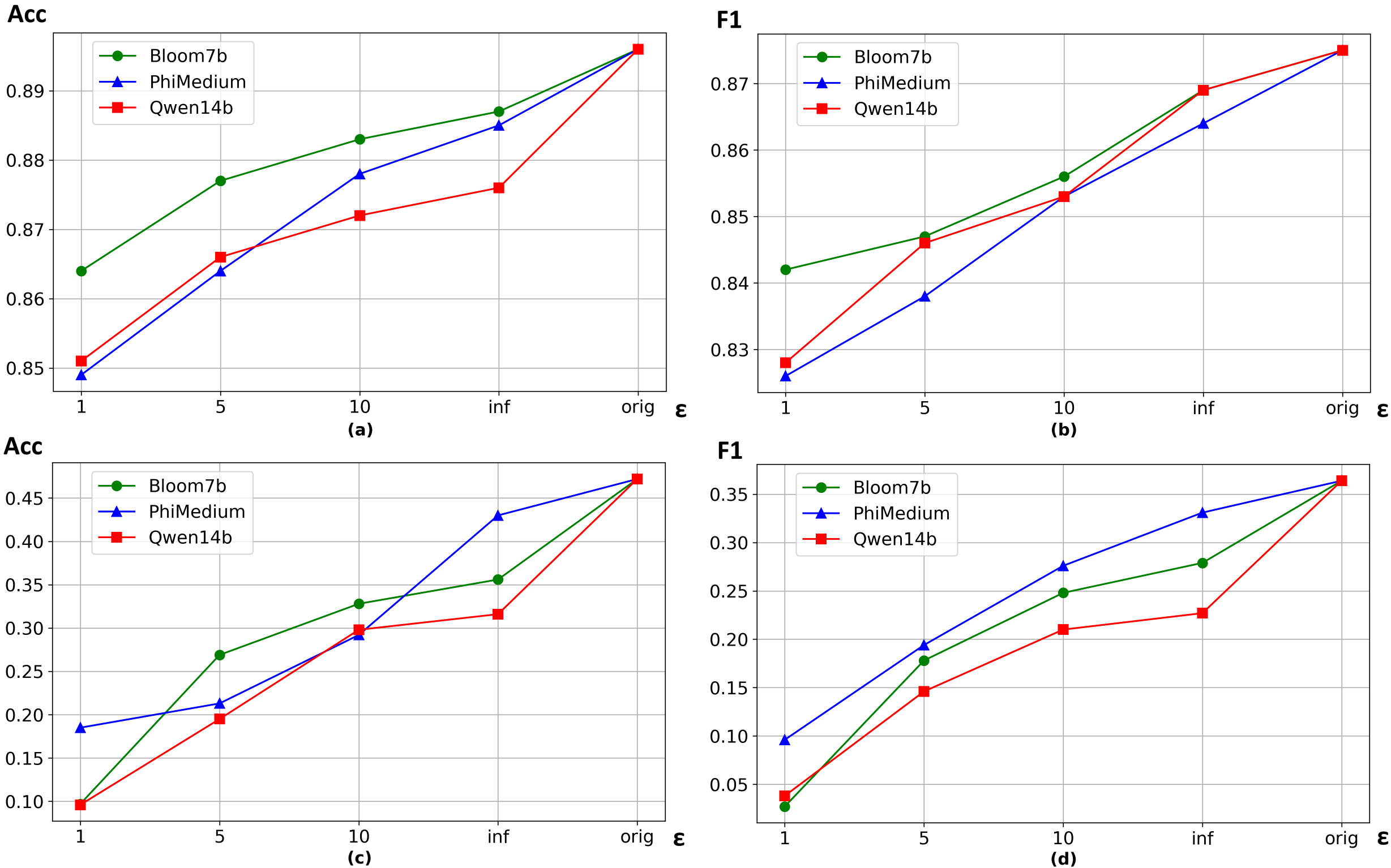}
    \caption{Classification accuracy and F1 scores for Autopsy on (a) and (b), and for Booksum on (c) and (d).}
    \label{fig:class}
\end{figure*}

One of the most common reasons for generating synthetic data is to use them for pretraining, training or tuning models for classification or similar tasks, especially in cases of natural data scarcity. As a result, benchmarking the private synthetic data on classification tasks is a good indicator of their quality and utility. 

We fine-tune ModernBERT-base with our synthetic data of various privacy levels to check how they compare against each other and against the original data on two classification tasks: (i) cause of death recognition based on Autopsy, and (ii) book genre recognition based on Booksum. The full details about these two tasks are presented in Appendix~\ref{sec:classtasks}. 

Classification accuracy and F$_1$ scores on cause of death recognition based on the synthetic Authopsy data are shown in Figure~\ref{fig:class} (a) and (b). As we can see, ModernBERT tuned with the Autopsy samples generated from Bloom7b, PhiMed, and Qwen14b yields accuracy scores ranging from 0.87, 0.84, and 0.85 for $\varepsilon = 1$ to 0.88, 0.88, and 0.87 for no privacy. The regression factors are 1.01, 1.05 and 1.02. Meanwhile, the accuracy score reached using the original Autopsy data is 0.89. 

The F$_1$ scores show a similar rising trend, ranging from 0.84, 0.83, and 0.83 for $\varepsilon = 1$ and rising up to 0.87, 0.86, and 0.87 for no privacy. They reveal regression factors of 1.04, 1.04, and 1.05. The F$_1$ score on the original Autopsy data is 0.87.

Booksum classification results in Figure~\ref{fig:class} (c) and (d) seem to be more sensitive to the privacy level. This has to do with the difficulty of the task which is about discriminating between 227 book genre categories. Accuracy starts at 0.09, 0.1, and 0.09 for the strongest privacy level, and rises up to 0.35, 0.4, and 0.31 for no privacy. The respective regression factors are 3.89, 4 and 3.44. The accuracy score when using the original Booksum data is 0.47. 

F$_1$ scores change very sharply, starting at 0.02, 0.02, and 0.03 and rising up to 0.27, 0.3, and 0.22. The respective regression factors are 13.5, 15, and 7.33. All the three scores are significantly lower than the F$_1$ score from the original Booksum data which is 0.36.

The classification scores on Autopsy are significantly higher than those on Booksum. We looped into the data and observed that one reason for that is the fact that each cause of death (e.g., ``Malaria'') which is the category to be predicted is verbally mentioned on the report in more than 95\,\% of the Autopsy samples. This probably helps ModernBERT to find the right category for the given verbal descriptions. Contrary, the genre (e.g., ``Adventure'') almost never appears in the book summaries. Another reason is the different numbers of categories that ModerBERT has to learn and recognize. In the case of Autopsy there are only 44 categories, but on Booksum data that grows to 227, increasing the classification difficulty.

\section{Discussion}
\label{sec:discussion}

In this work, we investigated the role that $(\varepsilon, \delta)$ DP plays in certain language quality aspects. We utilized two private datasets with sensitive data and a public dataset to fine-tune five generative LLMs under different privacy constrains and assessed different quality aspects of the texts they output. 

Our results reveal that stronger privacy constrains force LLMs to spill out shorter texts which implies a lower utility, since shorter texts usually convey less information. In fact, sample length sensitivity is significant, since output length decreases by at least 77\,\% and at most 494\,\% when the privacy budget changes from $\infty$ to 1.

Grammatical correctness of the synthetic samples is also penalized. We noticed grammatical accuracy reductions ranging from 9\,\% to 67\,\%. This happens in the form of additional word-level and sentence-level grammatical errors in the synthetic texts obtained from DP-tuned generative LLMs. 

Lexical diversity is another text quality aspect that is penalized. It usually comes in the form of more repeated words and n-grams within a generated text sample. When measured in terms of bi-gram diversity, the regression factor ranges from 10\,\% to 40\,\%. In terms of text compression ration, we observed increase levels from 8\,\% to 46\,\%.

Responding to \hyperlink{RQ1}{RQ1}, we conclude that strong DP has a strong negative impact on text length and a moderate negative impact on grammatical correctness and lexical diversity. 

We also observed the correlation between the privacy budget and the obtained quality scores. The difference between synthetic non-private outputs and the natural texts is negligible, but the one between the synthetic non-private outputs and the outputs of low privacy budgets is significant and gets bigger on lower privacy budget. 
In response to \hyperlink{RQ2}{RQ2}, we conclude that the smaller the privacy budget is, the lower the output length, the grammatical correctness, and the lexical diversity get. 

Besides output length, another form of text data utility is the performance we get from using them to tune LLMs on downstream tasks like classification. We observed slight classification accuracy and F$_1$ regressions on the task of cause of death recognition with Autopsy data, ranging from 1.01 (1\,\% lower) to 1.05 (5\,\% lower) for the accuracy and 1.02 (2\,\%lower) to 1.06 (6\,\% lower) for F$_1$. The impact on the task of book genre recognition with Booksum data is stronger, with regressions ranging from 3.44 (44\,\% lower) to 4 (300\,\% lower) for the accuracy and from 7.33 (633\,\% lower) to 15 (1400\,\% lower) for F$_1$. 

In response to \hyperlink{RQ3}{RQ3}, we conclude that the impact of DP tuning on the classification performance of the synthetic data it provides is negative and its sensitivity depends on the specific data and the tasks. This is in line with similar experimental results reported by \citet{yue_synthetic_2023}.  
It is worthy to note that there are other negative implications of DP which go beyond utility loss or language quality degradation. \citet{islam2024doesdifferentialprivacyimpact} show that DP-tuning can increase bias of the resulting LLMs against protected groups by making it more difficult to differentiate between positive and negative examples of those groups within a population. Furthermore, \citet{srivastava2024deamplifyingbiasdifferentialprivacy} investigate the trade-off between privacy and fairness and show that DP-tuned LLMs tend to be more biased. They also propose a method based on counterfactual data augmentation that helpt to de-amplify the bias that DP-tuning induces. 

All in all, the takeaway is that because of the various negative implications of DP-tuning in text utility, language quality, and other ethical aspects, a careful and comprehensive analysis is required in order to scale the adoption of DP-tuning for synthetic text generation.
An interesting future work could be to design a benchmarking suite that could help in standardizing the assessment of the tradeoff between privacy and the other quality aspects of the generated texts.

\section{Limitations}
\label{sec:limitations}

In this work, we used private and public data to examine the impact of LLM DP-tuning on language. 
We tuned several open LLMs with the original data and obtained synthetic data under different privacy constrains. A limitation of this work resulted from our restricted computation capacities. Unfortunately, the largest models that we managed to utilize do not exceed 14 billion parameters in size and are not the most performing ones today. Nevertheless, we believe that this limitation does not significantly impact the generalization of our results, since the five different LLMs we utilized represent diverse architectures and are pretrained with rich and huge text collections. 
Another limitation has to do with the evaluation part. Because of time constrains only automatic evaluation was performed. Adding human evaluation could provide some different and more versatile insights regarding the quality of the synthetic texts in general. 

\section{Ethical Considerations}
\label{sec:ethicalconsiderations}

In this work we used Autopsy and Suicide which are two private corpora with sensitive texts. In compliance with their licensing, we do not share the data with colleagues or other third parties and do not include them in the attachment that comes along with this submission. Instead, our experiments can be reproduced by obtaining the Authopsy\footnote{\url{https://osf.io/xuk5q/}} and the Suicide\footnote{\url{https://zenodo.org/records/6476179}} corpora directly from their authors, same as we did.  
Furthermore, we made sure to select and show in Tables~\ref{tab:autopsyexample} and \ref{tab:suicideexample} synthetic sample examples that do now reveal any sensitive details like location, cause of death, etc.

\section*{Acknowledgments}

This work has been partly supported by the Research Center Trustworthy Data
Science and Security (\url{https://rc-trust.ai}), one of the Research Alliance
centers within the \url{https://uaruhr.de}.

\bibliography{custom}

\appendix

\section{Model Hyperparameters}
\label{sec:tunedechyperparam}

We used the five models Bloom7b, PhiMini, PhiMed, Qwen7b, and Qwen14b with their default set of hyperparameters. The main model hyperparameters we adjusted during the fine-tuning phase are the number of epochs set to 6 for every model, and the batch size set to 16 for Bloom7b, PhiMini, and Qwen7b, and 8 for PhiMed and Qwen14b. The set of values for the adjusted model hyperparameters is shown in Table~\ref{tab:hyperparamqwen}. To optimize the memory, we fine-tuned each model using LoRA \cite{DBLP:journals/corr/abs-2106-09685}. The set of values for the adjusted LoRA hyperparameters is shown in Table~\ref{tab:hyperparam}.  

\begin{table}[ht]
  \centering
  \begin{tabular}{ l | l l l l }
    \hline
    \textbf{Model} & \textbf{Optim} & \textbf{LR} & \textbf{BS} & \textbf{NE} \\
    \hline
    Bloom7b & AdamW & $2 \cdot 10^{-5}$ & 16 & 6 \\
    Qween7b & AdamW & $2 \cdot 10^{-5}$ & 16 & 6 \\
    PhiMini & AdamW & $2 \cdot 10^{-5}$ & 16 & 6 \\
    Qwen14b & AdamW & $2 \cdot 10^{-5}$ & 8 & 6  \\
    PhiMed & AdamW & $2 \cdot 10^{-5}$ & 8 & 6 \\
    \hline
  \end{tabular}
  \caption{\label{tab:hyperparamqwen}Optimizer, learning rate (LR), batch size (BS), and number of epochs (NE) used to fine-tune the models.}
\end{table}

\begin{table}[t]
  \centering
  \begin{tabular}{ l | l }
    \hline
    \textbf{Hyperparam} & \textbf{Value} \\
    \hline
    low\_rank\_dim & 16 \\
    lora\_alpha & 32 \\
    lora\_dropout & 0.05 \\
    \hline
  \end{tabular}
  \caption{\label{tab:hyperparam}Lora hyper-parameters.}
\end{table}

\begin{table}[t]
\centering
\begin{tabular}{ l | r r }
    \hline
    \textbf{Part} & \textbf{Samples} & \textbf{Tokens} \\
    \hline
    Train & 6000 & 1227476 \\
    Test & 350 & 72848  \\
    Valid & 384 & 77946 \\
    Full & 6734 & 1378270  \\
    \hline
\end{tabular}
\caption{\label{tab:autopsystat}Autopsy statistics.}
\end{table}
\begin{table*}[t]
\begin{minipage}{.45\linewidth}
\centering
\begin{tabular}{ l | r r }
\hline
\textbf{Part} & \textbf{Samples} & \textbf{Tokens} \\
\hline
    Train & 10000 & 10442988 \\
    Test & 1400 & 1416321  \\
    Valid & 1441 & 1439347 \\
    Full & 12841 & 13298656 \\
\hline
\end{tabular}
\caption{\label{tab:booksumstat}Booksum statistics.}
\end{minipage}
\begin{minipage}{.45\linewidth}
\centering
\begin{tabular}{ l | r r }
    \hline
    \textbf{Part} & \textbf{Samples} & \textbf{Tokens} \\
    \hline
    Train & 14194 & 1226448 \\
    Test & 4427 & 381918  \\
    Valid & 3576 & 310921 \\
    Full & 22197 & 1919287  \\
    \hline
  \end{tabular}
  \caption{\label{tab:suicidestat}Suicide statistics.}
\end{minipage}
\end{table*}

\section{Corpora Statistics}
\label{sec:corporastats}

We turned each dataset into a file of JSON lines containing the data fields, and sliced each file in three parts: train, test, and valid. Their sizes in number of lines (samples) and number of tokens (for the text field) are presented in Tables~\ref{tab:autopsystat} -- \ref{tab:suicidestat}. 

\section{Classification Tasks}
\label{sec:classtasks}

To assess the classification utility of the synthetic texts produced with different privacy constrains we set up two classification tasks: cause of death recognition on the Autopsy data and book genre recognition on the Booksum data. For the first task, we fine-tuned multiple instances of ModernBERT with the different versions of the synthetic Autopsy data. In this case, the ModernBERT instances try to recognize the correct cause of death category from the total of 44 categories listed in Table~\ref{tab:autopsylabels}. For the second task, we fine-tuned multiple instances of ModernBERT with the different versions of the synthetic Booksum data. In this case, the ModernBERT instances try to recognize the correct book genre category from the total of 227 categories listed in Table~\ref{tab:booksumlabels}. We assessed the classification accuracy and the F$_1$ scores for each model instance and task.    

\begin{table*}[t]
\centering
\small
\begin{minipage}{0.24\textwidth}
\begin{tabular}{|l}
AIDS \\
Acute Myocardial Infarction \\
Asthma \\
Bite of Venomous Animal \\
Breast Cancer \\
COPD \\
Cervical Cancer \\
Cirrhosis \\
Colorectal Cancer \\
Diabetes \\
Diarrhea/Dysentery
\end{tabular}
\end{minipage} 
\begin{minipage}{0.24\textwidth}
\begin{tabular}{l}
Drowning \\
Encephalitis \\
Epilepsy \\
Esophageal Cancer \\
Falls \\
Fires \\
Hemorrhagic fever \\
Homicide \\
Leukemia/Lymphomas \\
Lung Cancer \\
Malaria
\end{tabular}
\end{minipage} 
\begin{minipage}{0.24\textwidth}
\begin{tabular}{l}
Maternal \\
Measles \\
Meningitis \\
Meningitis/Sepsis \\
Other Cancers \\
Other Cardiovascular \\
Other Child Deaths \\
Other Digestive \\
Other Infectious \\
Other Injuries \\
Other 
\end{tabular}
\end{minipage} 
\begin{minipage}{0.24\textwidth}
\begin{tabular}{l|}
Pneumonia \\
Poisonings \\
Prostate Cancer \\
Renal Failure \\
Road Traffic \\
Sepsis \\
Stomach Cancer \\
Stroke \\
Suicide \\
TB \\
Violent Death
\end{tabular}
\end{minipage} 
\caption{\label{tab:autopsylabels}List of the 44 cause of death labels in the Autopsy data.}
\end{table*}

\begin{table*}[t]
\small
\begin{minipage}{0.24\textwidth}
\begin{tabular}{|l}
Absurdist fiction \\
Adventure \\
Adventure novel \\
Albino bias \\
Alien invasion \\
Alternate history \\
American Gothic Fiction \\
Anthology \\
Anthropology \\
Anti-nuclear \\
Anti-war \\
Apocalyptic fiction \\
Autobiographical comics \\
Autobiographical novel \\
Autobiography \\
Bangsian fantasy \\
Bildungsroman \\
Biographical novel \\
Biography \\
Biopunk \\
Bit Lit \\
Black comedy \\
Boys' school stories \\
Business \\
Cabal \\
Campus novel \\
Catastrophic literature \\
Chick lit \\
Children's literature \\
Chivalric romance \\
Collage \\
Colonial US romance \\
Comedy \\
Comedy of manners \\
Comic book \\
Comic fantasy \\
Comic novel \\
Comic science fiction \\
Comics \\
Coming of age \\
Computer Science \\
Conspiracy \\
Conspiracy fiction \\
Contemporary fantasy \\
Cookbook \\
Cozy \\
Creative nonfiction \\
Crime Fiction \\
Cyberpunk \\
Dark fantasy \\
Detective fiction \\
Drama \\
Dying Earth subgenre \\
Dystopia \\
Economics \\
Edisonade \\
Education
\end{tabular}
\end{minipage}
\begin{minipage}{0.24\textwidth}
\begin{tabular}{l}
Elizabethan romance \\
Encyclopedia \\
English school stories \\
Epic Fantasy \\
Epistolary novel \\
Ergodic literature \\
Erotica \\
Essay \\
Existentialism \\
Experimental literature \\
Fable \\
Fairy tale \\
Fairytale fantasy \\
Fantastique \\
Fantasy \\
Fantasy of manners \\
Farce \\
Feminist science fiction \\
Fiction \\
Fictional crossover \\
Field guide \\
First-person narrative \\
Foreign legion \\
Future history \\
Gamebook \\
Gay Themed \\
Gay novel \\
Georgian romance \\
Ghost story \\
Gothic fiction \\
Graphic novel \\
Hard science fiction \\
Hardboiled \\
Heroic fantasy \\
High fantasy \\
Historical fantasy \\
Historical fiction \\
Historical novel \\
Historical romance \\
Historical whodunnit \\
History \\
Horror \\
Human extinction \\
Humour \\
Indian chick lit \\
Industrial novel \\
Inspirational \\
Invasion literature \\
Juvenile fantasy \\
Kunstlerroman \\
LGBT literature \\
Light novel \\
Literary criticism \\
Literary fiction \\
Literary realism \\
Literary theory \\
Locked room mystery
\end{tabular}
\end{minipage}
\begin{minipage}{0.24\textwidth}
\begin{tabular}{l}
Lost World \\
Low fantasy \\
Magic realism \\
Marketing \\
Mashup \\
Mathematics \\
Medieval romance \\
Memoir \\
Metaphysics \\
Military history \\
Military science fiction \\
Modernism \\
Morality play \\
Music \\
Mystery \\
Nature \\
Naval adventure \\
Neuroscience \\
New Weird \\
NY Times Best Seller \\
Non-fiction \\
Non-fiction novel \\
Novel \\
Novella \\
Parallel novel \\
Paranormal romance \\
Parody \\
Pastiche \\
Personal journal \\
Philosophy \\
Photography \\
Picaresque novel \\
Picture book \\
Planetary romance \\
Play \\
Poetry \\
Polemic \\
Police procedural \\
Political philosophy \\
Politics \\
Popular culture \\
Popular science \\
Pornography \\
Post-holocaust \\
Postcyberpunk \\
Postmodernism \\
Prose \\
Prose poetry \\
Psychological novel \\
Psychology \\
Reference \\
Regency romance \\
Religion \\
Religious text \\
Robinsonade \\
Role-playing game \\
Roman a clef
\end{tabular}
\end{minipage}
\begin{minipage}{0.24\textwidth}
\begin{tabular}{l|}
Romance novel \\
Romantic comedy \\
Satire \\
School story \\
Science \\
Science Fiction \\
Science fantasy \\
Scientific romance \\
Sea story \\
Self-help \\
Serial \\
Short story \\
Social commentary \\
Social criticism \\
Social novel \\
Social science fiction \\
Social sciences \\
Sociology \\
Soft science fiction \\
Space opera \\
Space western \\
Speculative fiction \\
Spirituality \\
Sports \\
Spy fiction \\
Steampunk \\
Subterranean fiction \\
Superhero fiction \\
Supernatural \\
Suspense \\
Sword and planet \\
Sword and sorcery \\
Techno-thriller \\
Thriller \\
Time travel \\
Tragicomedy \\
Transgender fiction \\
Transhumanism \\
Travel \\
Travel literature \\
Treatise \\
True crime \\
Urban fantasy \\
Urban fiction \\
Utopian fiction \\
Utopian fiction \\
Vampire fiction \\
War novel \\
Western \\
Western fiction \\
Whodunit \\
Wuxia \\
Young adult literature \\
Youth \\
Zombie \\
Zombies in culture
\end{tabular}
\end{minipage} 
\caption{\label{tab:booksumlabels}List of the 227 genre labels in the Booksum data.}
\end{table*}

\section{Examples of Synthetic Samples}
\label{sec:outputexamples}

Illustrations of the used prompts and the obtained text outputs for each dataset are illustrated in Tables~\ref{tab:autopsyexample} -- \ref{tab:suicideexample}. They were generated by PhiMed model with privacy budget $\varepsilon = 10$.

\begin{table*}[]
  \centering
  \begin{tabular}{| p{0.94\textwidth} |}
    \hline
    \ttfamily \textbf{Prompt:} Write a verbal authopsy of a female who was adult, 38 years old and died from Other Non-communicable Diseases: \vskip3mm
    \textbf{Output:} The autopsy report said she sustained blunt force injuries. She had bilateral post-mortem lividity. She seemed to have died from blunt force to the entire body, more prominent in the upper chest area. She also had multiple abrasions over the front. The blunt force trauma to the chest could have been caused by a stick. She did not have any injuries in the private parts, nor any defensive injuries. \\
    \hline
  \end{tabular}
  \caption{\label{tab:autopsyexample}Prompt and output example from PhiMed fine-tuned with Autopsy data and $\varepsilon = 10$.}
\end{table*}

\begin{table*}[]
  \centering
  \begin{tabular}{| p{0.94\textwidth} |}
    \hline
    \ttfamily \textbf{Prompt:} Write a plot summary for the book titled ``The Pedestrian'', written by Ray Bradbury of genre Science Fiction:  \vskip3mm
    \textbf{Output:} ``The Pedestrians'' by Ray Brabury is set in an unnamed city in Los Angeles, in the year 2052. At this time, leisure in cars has become so popularized for most people that pedestrianism has become an illegal activity. In this society, cars drive themselves, equipped with televisions, telephones, and food dispensers. People find this form of entertainment more gratifying and hence prefer it over any outdoor activities, and consequently, the streets are empty, except for the police patrols. The protagonist, Leonard Mead, is an individual who disobeys the norm and chooses the simple pleasure of taking long walks at night. He prefers to observe the dark, desolate world through his window, instead of the monotony of television. One evening, during one of his walks, he is stopped by a robotic police car. The officer explains that walking is illegal and, since cars are his only means of communication, he informs Mr. Mead's house about his arrest. However, the house seems to be uninhabited, except a few sleeping pills that fall from a medicine cabinet. The story concludes with the robotic police driving Leonard home, where he's left on the floor while the house, devoid. \\
    \hline
  \end{tabular}
  \caption{\label{tab:booksumexample}Prompt and output example from PhiMed fine-tuned with Booksum data and $\varepsilon = 10$.}
\end{table*}

\begin{table*}[]
  \centering
  \begin{tabular}{| p{0.94\textwidth} |}
    \hline
    \ttfamily \textbf{Prompt:} Write a post to the depression community:  \vskip3mm
    \textbf{Output:} Do you think therapy is worth it, especially during this pandemic? I've been really struggling to find a therapist who will meet my needs. And even when I do, the cost is just too much for me. It feels like the system is designed to keep us down in some ways. What's been your experience with therapy during these challenging times? Is it actually helpful, or should I be looking for different ways to cope? \\
    \hline
  \end{tabular}
  \caption{\label{tab:suicideexample}Prompt and output example from PhiMed fine-tuned with Suicide data and $\varepsilon = 10$.}
\end{table*}

\clearpage

\begin{table*}[h]
  \centering
  \begin{tabular}{l | r r r r r r r r}
    \hline
    \textbf{Configuration} & \textbf{NC} & \textbf{NT} & \textbf{WG} & \textbf{SG} & \textbf{UD} & \textbf{BD} & \textbf{HS} & \textbf{CR} \\
    \hline
    Natural, no DP & 980 & 180 & 0.977 & 0.694 & 0.093 & 0.427 & 0.114 & 3.295 \\
    Synthetic, no DP & 995 & 184 & 0.955 & 0.634 & 0.083 & 0.374 & 0.205 & 3.527 \\
    Synthetic, $\varepsilon = 10$ & 417 & 116 & 0.894 & 0.616 & 0.084 & 0.368 & 0.297 & 3.943 \\
    Synthetic, $\varepsilon = 5$ & 188 & 65 & 0.863 & 0.572 & 0.067 & 0.357 & 0.343 & 4.052 \\
    Synthetic, $\varepsilon = 1$ & 115 & 31 & 0.826 & 0.568 & 0.043 & 0.331 & 0.374 & 4.221 \\
    \hline
  \end{tabular}
  \caption{\label{tab:bloomautopsy}Scores of Bloom7b on Autopsy data.}
\end{table*}

\begin{table*}[h]
  \centering
  \begin{tabular}{l | r r r r r r r r}
    \hline
    \textbf{Configuration} & \textbf{NC} & \textbf{NT} & \textbf{WG} & \textbf{SG} & \textbf{UD} & \textbf{BD} & \textbf{HS} & \textbf{CR} \\
    \hline
    Natural, no DP & 2581 & 495 & 0.916 & 0.265 & 0.086 & 0.457 & 0.197 & 2.684 \\
    Synthetic, no DP & 2612 & 516 & 0.912 & 0.251 & 0.083 & 0.451 & 0.231 & 3.274 \\
    Synthetic, $\varepsilon = 10$ & 1526 & 304 & 0.896 & 0.188 & 0.075 & 0.397 & 0.294 & 3.882 \\
    Synthetic, $\varepsilon = 5$ & 1327 & 215 & 0.859 & 0.167 & 0.072 & 0.358 & 0.314 & 4.249 \\
    Synthetic, $\varepsilon = 1$ & 998 & 187 & 0.825 & 0.154 & 0.057 & 0.323 & 0.329 & 4.637 \\
    \hline
  \end{tabular}
  \caption{\label{tab:bloombooksum}Scores of Bloom7b on Booksum data.}
\end{table*}

\begin{table*}[h]
  \centering
  \begin{tabular}{l | r r r r r r r r}
    \hline
    \textbf{Configuration} & \textbf{NC} & \textbf{NT} & \textbf{WG} & \textbf{SG} & \textbf{UD} & \textbf{BD} & \textbf{HS} & \textbf{CR} \\
    \hline
    Natural, no DP & 718 & 142 & 0.965 & 0.624 & 0.095 & 0.335 & 0.25 & 2.753 \\
    Synthetic, no DP & 880 & 145 & 0.887 & 0.583 & 0.067 & 0.323 & 0.257 & 3.26 \\
    Synthetic, $\varepsilon = 10$ & 621 & 98 & 0.856 & 0.521 & 0.059 & 0.281 & 0.312 & 3.994 \\
    Synthetic, $\varepsilon = 5$ & 463 & 87 & 0.847 & 0.487 & 0.052 & 0.276 & 0.371 & 4.468 \\
    Synthetic, $\varepsilon = 1$ & 446 & 82 & 0.836 & 0.461 & 0.029 & 0.245 & 0.467 & 4.624 \\
    \hline
  \end{tabular}
  \caption{\label{tab:bloomsuicide}Scores of Bloom7b on Suicide data.}
\end{table*}

\begin{table*}[h]
  \centering
  \begin{tabular}{l | r r r r r r r r}
    \hline
    \textbf{Configuration} & \textbf{NC} & \textbf{NT} & \textbf{WG} & \textbf{SG} & \textbf{UD} & \textbf{BD} & \textbf{HS} & \textbf{CR} \\
    \hline
    Natural, no DP & 980 & 180 & 0.977 & 0.694 & 0.093 & 0.427 & 0.114 & 3.295 \\
    Synthetic, no DP & 967 & 177 & 0.963 & 0.692 & 0.094 & 0.415 & 0.264 & 3.768 \\
    Synthetic, $\varepsilon = 10$ & 590 & 123 & 0.921 & 0.624 & 0.072 & 0.365 & 0.316 & 4.055 \\
    Synthetic, $\varepsilon = 5$ & 385 & 96 & 0.885 & 0.587 & 0.052 & 0.354 & 0.329 & 4.341 \\
    Synthetic, $\varepsilon = 1$ & 279 & 71 & 0.853 & 0.575 & 0.044 & 0.327 & 0.356 & 4.419 \\
    \hline
  \end{tabular}
  \caption{\label{tab:aclb1b}Scores of Qwen7b on Autopsy data.}
\end{table*}

\begin{table*}[h]
  \centering
  \begin{tabular}{l | r r r r r r r r}
    \hline
    \textbf{Configuration} & \textbf{NC} & \textbf{NT} & \textbf{WG} & \textbf{SG} & \textbf{UD} & \textbf{BD} & \textbf{HS} & \textbf{CR} \\
    \hline
    Natural, no DP & 2581 & 495 & 0.916 & 0.265 & 0.086 & 0.457 & 0.197 & 2.684 \\
    Synthetic, no DP & 2614 & 514 & 0.912 & 0.264 & 0.065 & 0.426 & 0.243 & 3.675 \\
    Synthetic, $\varepsilon = 10$ & 1782 & 250 & 0.891 & 0.256 & 0.052 & 0.332 & 0.296 & 4.021 \\
    Synthetic, $\varepsilon = 5$ & 985 & 232 & 0.885 & 0.236 & 0.052 & 0.326 & 0.337 & 4.341 \\
    Synthetic, $\varepsilon = 1$ & 737 & 138 & 0.872 & 0.226 & 0.045 & 0.312 & 0.357 & 4.569 \\
    \hline
  \end{tabular}
  \caption{\label{tab:aclb1b}Scores of Qwen7b on Booksum data.}
\end{table*}

\begin{table*}[h]
  \centering
  \begin{tabular}{l | r r r r r r r r}
    \hline
    \textbf{Configuration} & \textbf{NC} & \textbf{NT} & \textbf{WG} & \textbf{SG} & \textbf{UD} & \textbf{BD} & \textbf{HS} & \textbf{CR} \\
    \hline
    Natural, no DP & 718 & 142 & 0.965 & 0.624 & 0.095 & 0.335 & 0.25 & 2.753 \\
    Synthetic, no DP & 724 & 163 & 0.93 & 0.603 & 0.077 & 0.306 & 0.243 & 3.134 \\
    Synthetic, $\varepsilon = 10$ & 619 & 126 & 0.915 & 0.542 & 0.034 & 0.262 & 0.296 & 4.221 \\
    Synthetic, $\varepsilon = 5$ & 520 & 109 & 0.882 & 0.494 & 0.023 & 0.221 & 0.352 & 4.476 \\
    Synthetic, $\varepsilon = 1$ & 474 & 86 & 0.832 & 0.434 & 0.017 & 0.192 & 0.385 & 4.665 \\
    \hline
  \end{tabular}
  \caption{\label{tab:aclb1b}Scores of Qwen7b on Suicide data.}
\end{table*}

\begin{table*}[h]
  \centering
  \begin{tabular}{l | r r r r r r r r}
    \hline
    \textbf{Configuration} & \textbf{NC} & \textbf{NT} & \textbf{WG} & \textbf{SG} & \textbf{UD} & \textbf{BD} & \textbf{HS} & \textbf{CR} \\
    \hline
    Natural, no DP & 980 & 180 & 0.977 & 0.694 & 0.093 & 0.427 & 0.114 & 3.295 \\
    Synthetic, no DP & 1035 & 203 & 0.971 & 0.683 & 0.092 & 0.416 & 0.251 & 3.388 \\
    Synthetic, $\varepsilon = 10$ & 655 & 134 & 0.951 & 0.643 & 0.086 & 0.383 & 0.286 & 3.756 \\
    Synthetic, $\varepsilon = 5$ & 570 & 118 & 0.943 & 0.635 & 0.062 & 0.377 & 0.313 & 3.827 \\
    Synthetic, $\varepsilon = 1$ & 496 & 105 & 0.916 & 0.605 & 0.057 & 0.372 & 0.328 & 3.946 \\
    \hline
  \end{tabular}
  \caption{\label{tab:aclb1b}Scores of Qwen14b on Autopsy data.}
\end{table*}

\begin{table*}[h]
  \centering
  \begin{tabular}{l | r r r r r r r r}
    \hline
    \textbf{Configuration} & \textbf{NC} & \textbf{NT} & \textbf{WG} & \textbf{SG} & \textbf{UD} & \textbf{BD} & \textbf{HS} & \textbf{CR} \\
    \hline
    Natural, no DP & 2581 & 495 & 0.916 & 0.265 & 0.086 & 0.457 & 0.197 & 2.684 \\
    Synthetic, no DP & 2080 & 410 & 0.907 & 0.225 & 0.068 & 0.442 & 0.215 & 2.991 \\
    Synthetic, $\varepsilon = 10$ & 1457 & 217 & 0.898 & 0.211 & 0.059 & 0.375 & 0.251 & 3.576 \\
    Synthetic, $\varepsilon = 5$ & 1017 & 203 & 0.893 & 0.194 & 0.052 & 0.356 & 0.268 & 3.883 \\
    Synthetic, $\varepsilon = 1$ & 979 & 196 & 0.857 & 0.168 & 0.048 & 0.343 & 0.289 & 4.364 \\
    \hline
  \end{tabular}
  \caption{\label{tab:aclb1b}Scores of Qwen14b on Booksum data.}
\end{table*}

\begin{table*}[h]
  \centering
  \begin{tabular}{l | r r r r r r r r}
    \hline
    \textbf{Configuration} & \textbf{NC} & \textbf{NT} & \textbf{WG} & \textbf{SG} & \textbf{UD} & \textbf{BD} & \textbf{HS} & \textbf{CR} \\
    \hline
    Natural, no DP & 718 & 142 & 0.965 & 0.624 & 0.095 & 0.335 & 0.25 & 2.753 \\
    Synthetic, no DP & 695 & 144 & 0.947 & 0.593 & 0.088 & 0.294 & 0.223 & 3.345 \\
    Synthetic, $\varepsilon = 10$ & 494 & 86 & 0.935 & 0.566 & 0.054 & 0.288 & 0.274 & 4.115 \\
    Synthetic, $\varepsilon = 5$ & 291 & 65 & 0.916 & 0.486 & 0.045 & 0.274 & 0.365 & 4.266 \\
    Synthetic, $\varepsilon = 1$ & 279 & 62 & 0.881 & 0.447 & 0.015 & 0.256 & 0.354 & 4.537 \\
    \hline
  \end{tabular}
  \caption{\label{tab:aclb1b}Scores of Qwen14b on Suicide data.}
\end{table*}

\begin{table*}[h]
  \centering
  \begin{tabular}{l | r r r r r r r r}
    \hline
    \textbf{Configuration} & \textbf{NC} & \textbf{NT} & \textbf{WG} & \textbf{SG} & \textbf{UD} & \textbf{BD} & \textbf{HS} & \textbf{CR} \\
    \hline
    Natural, no DP & 980 & 180 & 0.977 & 0.694 & 0.093 & 0.427 & 0.114 & 3.295 \\
    Synthetic, no DP & 997 & 194 & 0.954 & 0.627 & 0.094 & 0.416 & 0.195 & 3.426 \\
    Synthetic, $\varepsilon = 10$ & 671 & 132 & 0.915 & 0.582 & 0.084 & 0.396 & 0.394 & 3.577 \\
    Synthetic, $\varepsilon = 5$ & 446 & 86 & 0.945 & 0.536 & 0.076 & 0.394 & 0.466 & 3.644 \\
    Synthetic, $\varepsilon = 1$ & 397 & 73 & 0.847 & 0.494 & 0.058 & 0.374 & 0.525 & 3.814 \\
    \hline
  \end{tabular}
  \caption{\label{tab:aclb1b}Scores of Phi mini on Autopsy data.}
\end{table*}

\begin{table*}[h]
  \centering
  \begin{tabular}{l | r r r r r r r r}
    \hline
    \textbf{Configuration} & \textbf{NC} & \textbf{NT} & \textbf{WG} & \textbf{SG} & \textbf{UD} & \textbf{BD} & \textbf{HS} & \textbf{CR} \\
    \hline
    Natural, no DP & 2581 & 495 & 0.916 & 0.265 & 0.086 & 0.457 & 0.197 & 2.684 \\
    Synthetic, no DP & 2578 & 489 & 0.906 & 0.234 & 0.062 & 0.421 & 0.24 & 3.594 \\
    Synthetic, $\varepsilon = 10$ & 1411 & 298 & 0.881 & 0.224 & 0.057 & 0.357 & 0.28 & 3.964 \\
    Synthetic, $\varepsilon = 5$ & 1223 & 221 & 0.834 & 0.186 & 0.054 & 0.348 & 0.312 & 4.145 \\
    Synthetic, $\varepsilon = 1$ & 1140 & 202 & 0.824 & 0.156 & 0.043 & 0.317 & 0.355 & 4.628 \\
    \hline
  \end{tabular}
  \caption{\label{tab:aclb1b}Scores of Phi mini on Booksum data.}
\end{table*}

\begin{table*}[h]
  \centering
  \begin{tabular}{l | r r r r r r r r}
    \hline
    \textbf{Configuration} & \textbf{NC} & \textbf{NT} & \textbf{WG} & \textbf{SG} & \textbf{UD} & \textbf{BD} & \textbf{HS} & \textbf{CR} \\
    \hline
    Natural, no DP & 718 & 142 & 0.965 & 0.624 & 0.095 & 0.335 & 0.25 & 2.753 \\
    Synthetic, no DP & 734 & 164 & 0.943 & 0.625 & 0.087 & 0.295 & 0.226 & 3.625 \\
    Synthetic, $\varepsilon = 10$ & 573 & 126 & 0.914 & 0.566 & 0.045 & 0.367 & 0.314 & 4.247 \\
    Synthetic, $\varepsilon = 5$ & 449 & 101 & 0.883 & 0.537 & 0.044 & 0.246 & 0.343 & 4.55 \\
    Synthetic, $\varepsilon = 1$ & 385 & 83 & 0.857 & 0.484 & 0.031 & 0.217 & 0.356 & 4.765 \\
    \hline
  \end{tabular}
  \caption{\label{tab:aclb1b}Scores of Phi mini on Suicide data.}
\end{table*}

\begin{table*}[h]
  \centering
  \begin{tabular}{l | r r r r r r r r}
    \hline
    \textbf{Configuration} & \textbf{NC} & \textbf{NT} & \textbf{WG} & \textbf{SG} & \textbf{UD} & \textbf{BD} & \textbf{HS} & \textbf{CR} \\
    \hline
    Natural, no DP & 980 & 180 & 0.977 & 0.694 & 0.093 & 0.427 & 0.114 & 3.295 \\
    Synthetic, no DP & 998 & 193 & 0.965 & 0.607 & 0.094 & 0.416 & 0.278 & 3.423 \\
    Synthetic, $\varepsilon = 10$ & 592 & 114 & 0.926 & 0.583 & 0.084 & 0.387 & 0.474 & 3.496 \\
    Synthetic, $\varepsilon = 5$ & 375 & 71 & 0.913 & 0.568 & 0.085 & 0.386 & 0.484 & 3.532 \\
    Synthetic, $\varepsilon = 1$ & 359 & 68 & 0.91 & 0.517 & 0.066 & 0.354 & 0.56 & 3.634 \\
    \hline
  \end{tabular}
  \caption{\label{tab:aclb1b}Scores of Phi medium on Autopsy data.}
\end{table*}

\begin{table*}[h]
  \centering
  \begin{tabular}{l | r r r r r r r r}
    \hline
    \textbf{Configuration} & \textbf{NC} & \textbf{NT} & \textbf{WG} & \textbf{SG} & \textbf{UD} & \textbf{BD} & \textbf{HS} & \textbf{CR} \\
    \hline
    Natural, no DP & 2581 & 495 & 0.916 & 0.265 & 0.086 & 0.457 & 0.197 & 2.684 \\
    Synthetic, no DP & 2655 & 512 & 0.895 & 0.234 & 0.087 & 0.435 & 0.197 & 3.144 \\
    Synthetic, $\varepsilon = 10$ & 1399 & 268 & 0.866 & 0.168 & 0.073 & 0.357 & 0.223 & 3.568 \\
    Synthetic, $\varepsilon = 5$ & 902 & 195 & 0.847 & 0.167 & 0.056 & 0.347 & 0.243 & 4.071 \\
    Synthetic, $\varepsilon = 1$ & 878 & 164 & 0.845 & 0.155 & 0.05 & 0.315 & 0.284 & 4.348 \\
    \hline
  \end{tabular}
  \caption{\label{tab:aclb1b}Scores of Phi medium on Booksum data.}
\end{table*}

\begin{table*}[h]
  \centering
  \begin{tabular}{l | r r r r r r r r}
    \hline
    \textbf{Configuration} & \textbf{NC} & \textbf{NT} & \textbf{WG} & \textbf{SG} & \textbf{UD} & \textbf{BD} & \textbf{HS} & \textbf{CR} \\
    \hline
    Natural, no DP & 718 & 142 & 0.965 & 0.624 & 0.095 & 0.335 & 0.25 & 2.753 \\
    Synthetic, no DP & 776 & 153 & 0.954 & 0.583 & 0.066 & 0.313 & 0.213 & 3.515 \\
    Synthetic, $\varepsilon = 10$ & 742 & 148 & 0.944 & 0.576 & 0.045 & 0.293 & 0.285 & 4.125 \\
    Synthetic, $\varepsilon = 5$ & 492 & 109 & 0.945 & 0.565 & 0.033 & 0.287 & 0.293 & 4.36 \\
    Synthetic, $\varepsilon = 1$ & 385 & 84 & 0.845 & 0.533 & 0.024 & 0.265 & 0.343 & 4.527 \\
    \hline
  \end{tabular}
  \caption{\label{tab:phimediumsuicide}Scores of Phi medium on Suicide data.}
\end{table*}

\begin{table}[h]
  \centering
  \begin{tabular}{l | r r}
    \hline
    \textbf{Configuration} & \textbf{Acc} & \textbf{F\textsubscript{1}} \\
    \hline
    Natural, no DP & 0.896 & 0.875 \\
    Synthetic, no DP & 0.885 & 0.864 \\
    Synthetic, $\varepsilon = 10$ & 0.878 & 0.853 \\
    Synthetic, $\varepsilon = 5$ & 0.864 & 0.838 \\
    Synthetic, $\varepsilon = 1$ & 0.849 & 0.826 \\
    \hline
  \end{tabular}
  \caption{\label{tab:phimediumautopsyclass}Phi medium Autopsy classification.}
\end{table}

\begin{table}[h]
  \centering
  \begin{tabular}{l | r r}
    \hline
    \textbf{Configuration} & \textbf{Acc} & \textbf{F\textsubscript{1}} \\
    \hline
    Natural, no DP & 0.472 & 0.364 \\
    Synthetic, no DP & 0.43 & 0.331 \\
    Synthetic, $\varepsilon = 10$ & 0.292 & 0.276 \\
    Synthetic, $\varepsilon = 5$ & 0.213 & 0.194 \\
    Synthetic, $\varepsilon = 1$ & 0.185 & 0.096 \\
    \hline
  \end{tabular}
  \caption{\label{tab:phimediumbooksumclass}Phi medium Booksum classification.}
\end{table}

\begin{table}[h]
  \centering
  \begin{tabular}{l | r r}
    \hline
    \textbf{Configuration} & \textbf{Acc} & \textbf{F\textsubscript{1}} \\
    \hline
    Natural, no DP & 0.896 & 0.875 \\
    Synthetic, no DP & 0.882 & 0.86 \\
    Synthetic, $\varepsilon = 10$ & 0.875 & 0.856 \\
    Synthetic, $\varepsilon = 5$ & 0.867 & 0.844 \\
    Synthetic, $\varepsilon = 1$ & 0.858 & 0.829 \\
    \hline
  \end{tabular}
  \caption{\label{tab:phiminiautopsyclass}Phi mini Autopsy classification.}
\end{table}

\begin{table}[h]
  \centering
  \begin{tabular}{l | r r}
    \hline
    \textbf{Configuration} & \textbf{Acc} & \textbf{F\textsubscript{1}} \\
    \hline
    Natural, no DP & 0.472 & 0.364 \\
    Synthetic, no DP & 0.342 & 0.194 \\
    Synthetic, $\varepsilon = 10$ & 0.29 & 0.153 \\
    Synthetic, $\varepsilon = 5$ & 0.215 & 0.096 \\
    Synthetic, $\varepsilon = 1$ & 0.197 & 0.047 \\
    \hline
  \end{tabular}
  \caption{\label{tab:phiminibooksumclass}Phi mini Booksum classification.}
\end{table}

\begin{table}[h]
  \centering
  \begin{tabular}{l | r r}
    \hline
    \textbf{Configuration} & \textbf{Acc} & \textbf{F\textsubscript{1}} \\
    \hline
    Natural, no DP & 0.896 & 0.875 \\
    Synthetic, no DP & 0.887 & 0.869 \\
    Synthetic, $\varepsilon = 10$ & 0.883 & 0.856 \\
    Synthetic, $\varepsilon = 5$ & 0.877 & 0.847 \\
    Synthetic, $\varepsilon = 1$ & 0.864 & 0.842 \\
    \hline
  \end{tabular}
  \caption{\label{tab:bloom7autopsyclass}Bloom7b Autopsy classification.}
\end{table}

\begin{table}[h]
  \centering
  \begin{tabular}{l | r r}
    \hline
    \textbf{Configuration} & \textbf{Acc} & \textbf{F\textsubscript{1}} \\
    \hline
    Natural, no DP & 0.472 & 0.364 \\
    Synthetic, no DP & 0.356 & 0.279 \\
    Synthetic, $\varepsilon = 10$ & 0.328 & 0.248 \\
    Synthetic, $\varepsilon = 5$ & 0.269 & 0.178 \\
    Synthetic, $\varepsilon = 1$ & 0.097 & 0.027 \\
    \hline
  \end{tabular}
  \caption{\label{tab:bloom7booksumclass}Bloom7b Booksum classification.}
\end{table}

\begin{table}[h]
  \centering
  \begin{tabular}{l | r r}
    \hline
    \textbf{Configuration} & \textbf{Acc} & \textbf{F\textsubscript{1}} \\
    \hline
    Natural, no DP & 0.896 & 0.875 \\
    Synthetic, no DP & 0.884 & 0.865 \\
    Synthetic, $\varepsilon = 10$ & 0.877 & 0.846 \\
    Synthetic, $\varepsilon = 5$ & 0.862 & 0.815 \\
    Synthetic, $\varepsilon = 1$ & 0.839 & 0.797 \\
    \hline
  \end{tabular}
  \caption{\label{tab:qwen7autopsyclass}Qwen7b Autopsy classification.}
\end{table}

\begin{table}[h]
  \centering
  \begin{tabular}{l | r r}
    \hline
    \textbf{Configuration} & \textbf{Acc} & \textbf{F\textsubscript{1}} \\
    \hline
    Natural, no DP & 0.472 & 0.364 \\
    Synthetic, no DP & 0.375 & 0.26 \\
    Synthetic, $\varepsilon = 10$ & 0.288 & 0.208 \\
    Synthetic, $\varepsilon = 5$ & 0.236 & 0.146 \\
    Synthetic, $\varepsilon = 1$ & 0.118 & 0.048 \\
    \hline
  \end{tabular}
  \caption{\label{tab:qwen7booksumclass}Qwen7b Booksum classification.}
\end{table}

\begin{table}[h]
  \centering
  \begin{tabular}{l | r r}
    \hline
    \textbf{Configuration} & \textbf{Acc} & \textbf{F\textsubscript{1}} \\
    \hline
    Natural, no DP & 0.896 & 0.875 \\
    Synthetic, no DP & 0.876 & 0.869 \\
    Synthetic, $\varepsilon = 10$ & 0.872 & 0.853 \\
    Synthetic, $\varepsilon = 5$ & 0.866 & 0.846 \\
    Synthetic, $\varepsilon = 1$ & 0.851 & 0.828 \\
    \hline
  \end{tabular}
  \caption{\label{tab:qwen14autopsyclass}Qwen14b Autopsy classification.}
\end{table}

\begin{table}[h]
  \centering
  \begin{tabular}{l | r r}
    \hline
    \textbf{Configuration} & \textbf{Acc} & \textbf{F\textsubscript{1}} \\
    \hline
    Natural, no DP & 0.472 & 0.364 \\
    Synthetic, no DP & 0.316 & 0.227 \\
    Synthetic, $\varepsilon = 10$ & 0.298 & 0.21 \\
    Synthetic, $\varepsilon = 5$ & 0.195 & 0.146 \\
    Synthetic, $\varepsilon = 1$ & 0.096 & 0.038 \\
    \hline
  \end{tabular}
  \caption{\label{tab:qwen14booksumclass}Qwen14b Booksum classification.}
\end{table}

\clearpage

\section{Full Result Scores}
\label{sec:fullscores}

The full language quality scores of each model on each dataset are shown in Tables~\ref{tab:bloomautopsy} -- \ref{tab:phimediumsuicide}. We show each of the scores for the eight evaluation metrics discussed in \S~\ref{ssec:evalmetrics}. The scores were computed by each combination of the five models tuned with each of the three datasets for each of the four privacy levels. The first line of each table shows the scores computed for the original natural texts of each dataset. 
Tables~\ref{tab:phimediumautopsyclass} -- \ref{tab:qwen14booksumclass} show the accuracy and F$_1$ scores of each model fine-tuned with the Autopsy and Booksum data on the cause of death recognition and genre recognition classification tasks. 

\end{document}